\documentclass{article}

\PassOptionsToPackage{options}{natbib}
\usepackage[final,nonatbib]{neurips_2021}

\usepackage[utf8]{inputenc} 
\usepackage[T1]{fontenc}    
\usepackage{hyperref}       
\usepackage{url}            
\usepackage{booktabs}       
\usepackage{amsfonts}       
\usepackage{nicefrac}       
\usepackage{microtype}      
\usepackage{xcolor}         

\title{An Empirical Study of Finding Similar Exercises}

\author{%
  Tongwen Huang, Xihua Li \\
  Tencent Cloud Xiaowei, China\\
  \texttt{\{towanhuang,xihuali\}@tencent.com} \\
}

\usepackage{enumerate}
\usepackage{multirow}
\usepackage[pdftex]{graphicx}
\usepackage{subfigure}

\usepackage{amsthm}       
\usepackage{CJKutf8}

\usepackage{caption}
\usepackage{wrapfig}

\begin{document}

\maketitle

\begin{abstract}
Education artificial intelligence aims to profit tasks in the education domain such as intelligent test paper generation and consolidation exercises where the main technique behind is how to match the exercises, known as the finding similar exercises(FSE) problem. 
Most of these approaches emphasized their model abilities to represent the exercise, unfortunately there are still many challenges such as the scarcity of data, insufficient understanding of exercises and high label noises. We release a Chinese education pre-trained language model BERT$_{Edu}$ for the label-scarce dataset and introduce the exercise normalization to overcome the diversity of mathematical formulas and terms in exercise. We discover new auxiliary tasks in an innovative way depends on problem-solving ideas and propose a very effective MoE enhanced multi-task model for FSE task to attain better understanding of exercises. In addition, confidence learning was utilized to prune train-set and overcome high noises in labeling data. Experiments show that these methods proposed in this paper are very effective.
\end{abstract}

\section{Introduction}
\label{s1}

With the rapid informatization of education, education artificial intelligence(EduAI) focuses on applying methods of artificial intelligence to benefit education tasks, which can help to improve the performance of the students and the teaching quality of the teachers. Intelligent test paper generation and consolidation exercises are important exercise-based applications. The teachers choose the similar exercises to replace their not satisfied ones when they generate the test paper and the students obtain some personalized consolidation exercises to improve their performances. The main technique behind is how to match the exercises, known as the finding similar exercises(FSE) problem.

\begin{figure*}[!hbt]
  \centering
\includegraphics[width=0.95\linewidth]{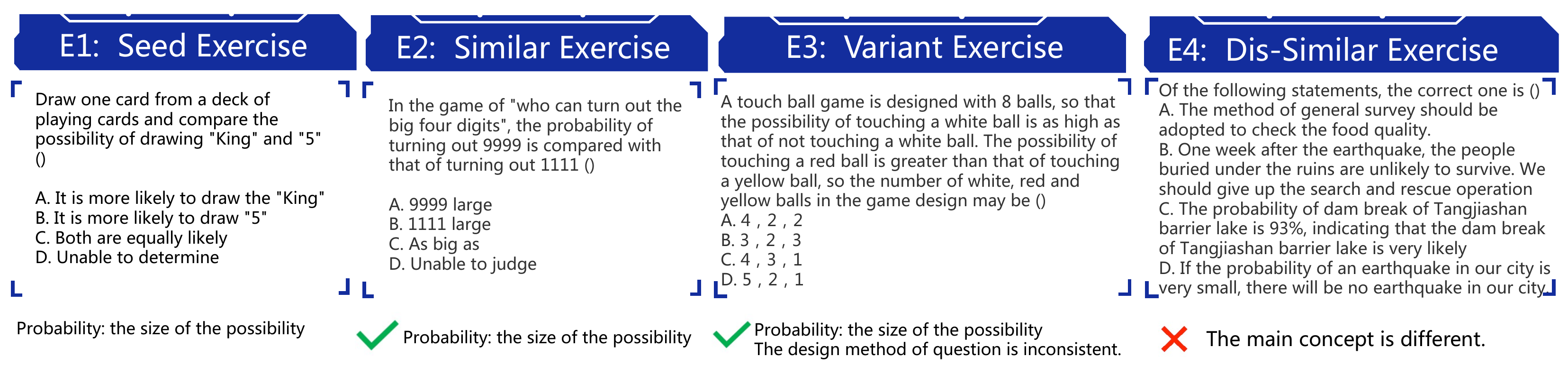}
  \caption{Examples of math exercises: E1, E2, E3 and E4.}
  \label{fig:fmath}
\end{figure*}

The generalized similar exercise includes \textit{similar exercise} and \textit{variant exercise}, which is concerned with finding an exercise has a similar logic and relationships by understanding the semantics of exercise. As example in Figure-\ref{fig:fmath}, where exercise E1 and its similar ones E2 and E3 share the same purpose of assessing the size of the possibility, while the dissimilar exercises E1 and E4 have different purposes. From above, we find that it is hard to understand the semantics of exercises from contents only. Some prior works leveraged the texts and concepts of exercises to calculate exercise similarity. For example, Vector Space Model(VSM)\cite{pelanek2019measuring} represents the exercises with TF-IDF and measures the similarity of exercises with text distance methods. In recent years, many deep learning based methods have been proposed and achieved success such as MANN\cite{liu2018finding}, SBERT\cite{feng2019learning}, QuesNet\cite{yin2019quesnet} etc.

However, there are still many challenges that paid less attention in EduAI such as the scarcity of data, high label noises and insufficient semantically understanding of exercises. First, due to the privately-owned exercise bank, it is difficult for us to have a general pre-training model in educational knowledge like BERT in NLP\cite{devlin2018bert}. Second, the human labeled educational data requires expertise and much time, which is scarce and expensive, so how to effectively utilize limit data? Third, owing to the difficulty and easy-confused of similar exercises, there is a lot of label noises, how to learn a model in noise dataset is a big challenge. Fourth, because of the multiple structural attributes of the exercise, how to obtain a better logical and semantic representation by its structure?

To address the challenges mentioned above, a new model named ExerciseBERT was proposed for FSE in this paper. The main features of ExerciseBERT are concluded as follows:

\begin{itemize}

\item We release a Chinese education pre-trained language model BERT$_{Edu}$ and boost the performance in the label-scarce dataset for downstream tasks;
\item For the diversity of mathematical formulas and terms in exercise, we introduce the exercise normalization method to ensure its identity;
\item We discover new auxiliary tasks in an innovative way depends on problem-solving ideas and propose a very effective MoE enhanced multi-task model for FSE task to attain better understanding of exercises;
\item In addition, we utilize confidence learning to prune train-set and overcome noise in data.
\end{itemize}


\section{Related Work}
\label{s2}


Early works attempt to analyze the text similarity based on vector space
model, where the same concepts or the similar words are used to calculate exercise similarity. Yu\cite{yu2014similarity} proposed a text match method by knowledge tree and keywords matching. \cite{pelanek2019measuring} utilized various unsupervised distance measurement methods and Kappa learning \cite{nazaretsky2018kappa} was an improved method for Kappa distance. In recent years, semantic representations/understandings have made a breakthrough. Test-aware Attention-based Convolutional Neural Network
(TACNN) is a model that utilizes the semantic representations of text materials(document, question and options) to predict exam question difficulty. MANN\cite{liu2018finding} utilizes the Long Short-Term Memory (LSTM) model to capture the exercise representation, targeting at the FSE task while SBERT\cite{feng2019learning} utilizes the transformer to encode the exercise and improves model accuracy. QuesNet\cite{yin2019quesnet} is a question embedding model pre-trained and introduces another task called Domain-Oriented Objective (DOO) to capture high-level logical information.
Most of these approaches emphasize their model abilities to represent the exercise, unfortunately there still many challenges in FSE as mentioned above. 

\section{Challenges in FSE}
\label{s4}

\begin{figure}[!ht]
  \centering
  \includegraphics[height=0.26\linewidth]{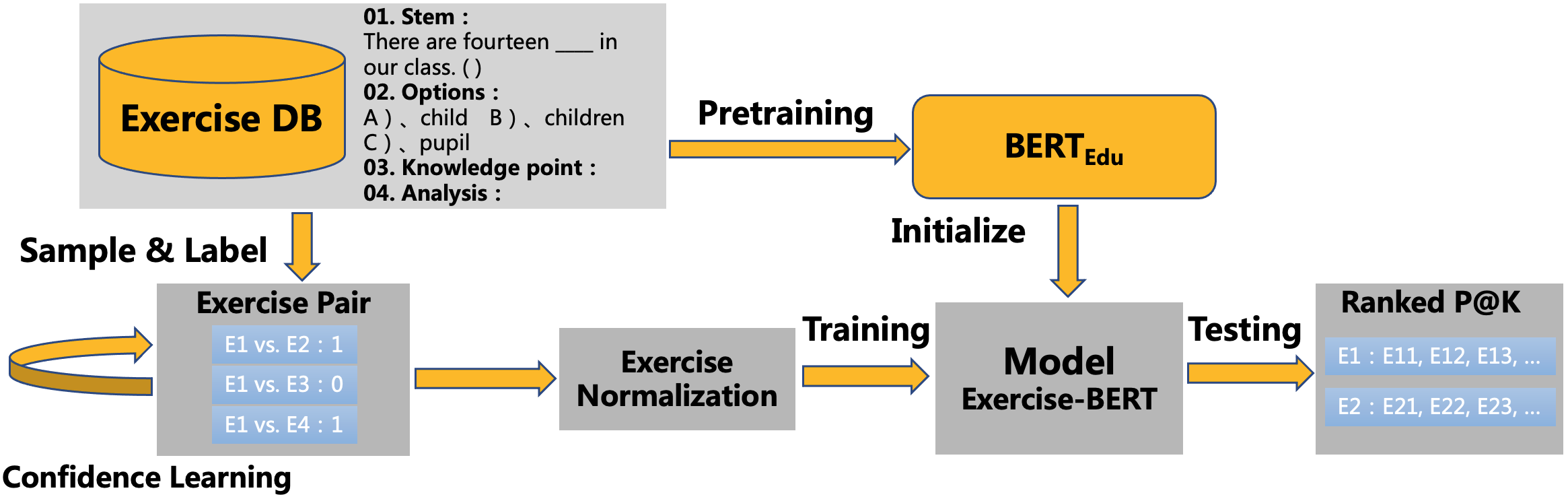}
    \caption{The overall workflow of ExerciseBERT for FSE tasks. Pairs rated as “similar question” by humans were labeled “1”, pairs rated as “dis-similar” were labeled “0”.}
  \label{fig:f1}
\end{figure}
As mentioned earlier, similar exercises are those having the same purpose which is related with the semantics of exercises. The formal mathematical definition is described in the Appendix-\ref{s30}. For tackling with these challenges, the overview of our solution shows in the Figure-\ref{fig:f1}. We'll go through each stage with the corresponding problems.


\subsection{Pre-training: Limits Data and Lack of Education Knowledge for Challenge 1}
\label{s40}

The annotation of similar exercises is very expensive, scarce and time-consuming because of its strong expertise, while millions of un-labeled exercises contain abundant domain logic and knowledge. Inspired by the effectiveness of pre-training models, which trained on amounts of unlabelled corpora, are able to benefit a variety of downstream NLP tasks. We also pre-train the data in the field of education. The pre-training dataset consists of a large education domain corpus of all structural exercise database. The pre-training language task objective is BERT-WWM\cite{chinese-bert-wwm} which adopts whole word masking rather than single character masking for pre-training BERT and obtains a good performance among Chinese pre-trained models.

\subsection{Exercise Normalization: the Diversity of Exercise for Challenge 2}
\label{s400}
Taking the E as the example in Table \ref{table:t400}, E1 is the original concise exercise, while E2 and E3 include the rendering of styles such as `mathrm' and `textit'. 


\begin{table}[!htbp]
    \caption{The Diversity of Exercise Representations}
    \label{table:t400}
    \centering
    \scriptsize
    \begin{tabular}{ll}
    \hline
         & Example \\ \toprule
        E & If  $\sqrt[3x-10]{2x+y-5}$ and $ \sqrt{x-3y+11}$ are homogeneous quadratic radicals, find the values of x and y. \\ \hline
        E1 & If  \$ $\backslash$sqrt[3x-10]\{2x+y-5\}\$ and \$ $\backslash$sqrt\{x-3y+11\}\$  are homogeneous quadratic radicals,find the values of x and y. \\ \hline
        E2 & If \$ $\backslash$sqrt[3x-$\backslash$text\{-\}10]\{2x+y$\backslash$text\{-\}5\}\$ and \$ $\backslash$sqrt\{x $\backslash$text\{-\}3y+11\}\$ are homogeneous quadratic radicals, find the values \\ & of x and y. \\ \hline
        E3 & If \$ $\backslash$sqrt[3x$\backslash$text\{-\}10]\{2$\backslash$mathrm\{x\}+$\backslash$mathrm\{y\}  $\backslash$text\{-\}5\}\$ and \$ $\backslash$sqrt\{$\backslash$mathrm\{x\}$\backslash$text\{-\}3 $\backslash$mathrm\{y\}+11\}\$ are \\ & homogeneous quadratic radicals, find the values of x and y. \\  \bottomrule
    \end{tabular}
\end{table}
From the perspective of keywords matching, E1, E2 and E3 are very different but they share the same meanings as E, so we need the \textit{exercise normalization}. We utilize the syntax parsing tool ANTLR\cite{antlr4} to norm the mathematical formula in the exercise. Taking the above example, the ``\$ $\backslash$sqrt[3x-10]\{2x+y-5\}\$'', ``\$ $\backslash$sqrt[3x-$\backslash$text\{-\}10]\{2x+y$\backslash$text\{-\}5\}\$'' and ``\$ $\backslash$sqrt[3x$\backslash$text\{-\}10]\{2$\backslash$mathrm\{x\}+$\backslash$mathrm\{y\}  $\backslash$text\{-\}5\}\$'' can be normalized the same form `root(2x+y-5, 3x-10)' to ensure identity. Similarly, technical terms, HTML or CSS tag also need to be normalized.

\subsection{Multi-Task Learning: Insufficient Semantically Understanding for Challenge 3}
\label{s42}
Normal semantic understanding of similar exercises, which only depends on stems and options is not enough. In fact, the judgment of similar exercises not only needs to understand the exercises, but also needs to know how to solve the exercises. Therefore, how to assist the judgment of similar exercises with the help of problem-solving ideas?

\subsubsection{Discovering the New Tasks}
\label{s43}
\begin{figure}[ht]

\begin{minipage}[t]{0.5\linewidth}
\centering
\includegraphics[height=0.48\linewidth]{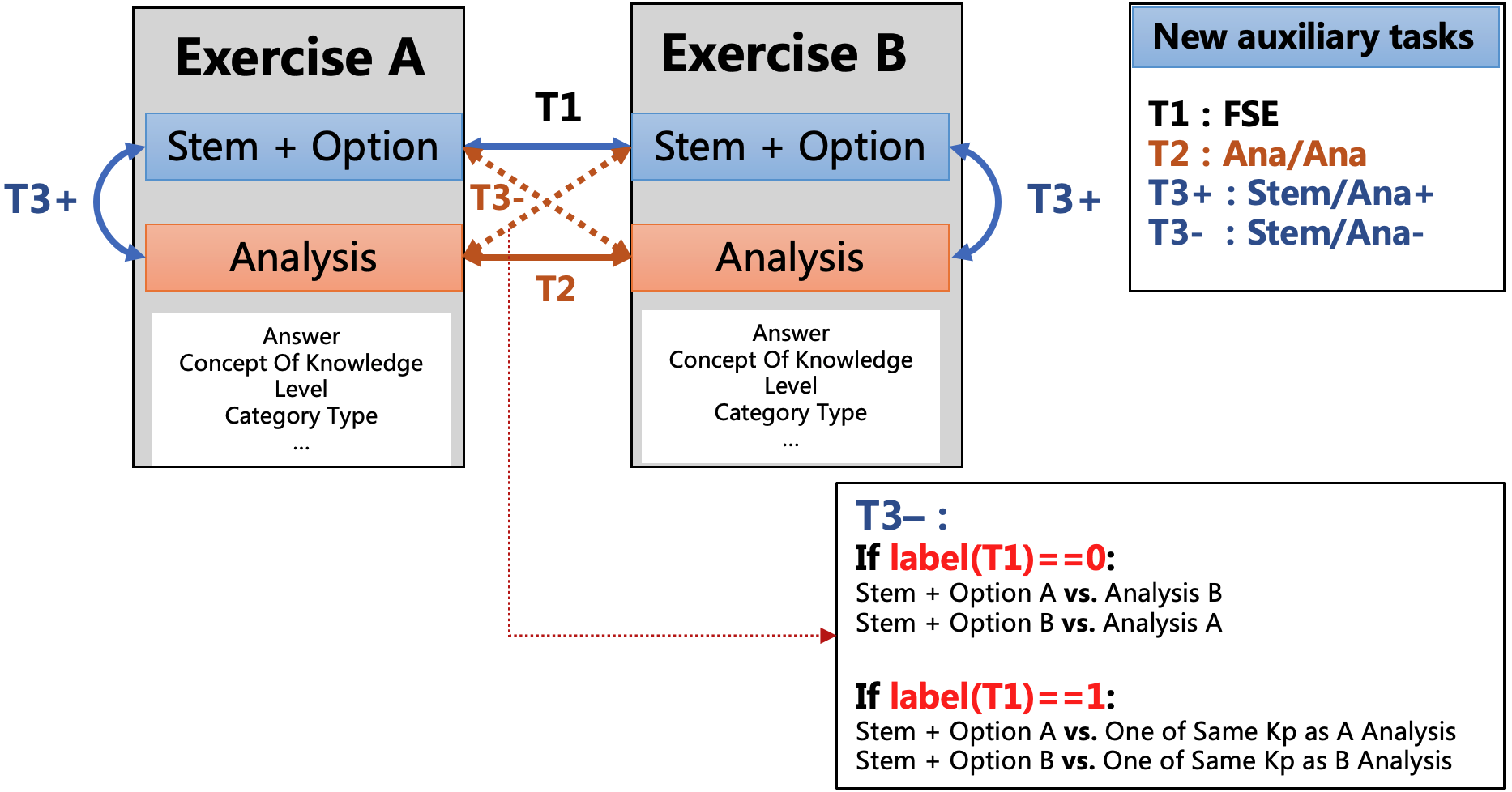}
\caption{The new auxiliary tasks. }
\label{fig:f4}
\end{minipage}
\hfill 
\begin{minipage}[t]{0.5\linewidth}
\centering
\includegraphics[height=0.48\linewidth]{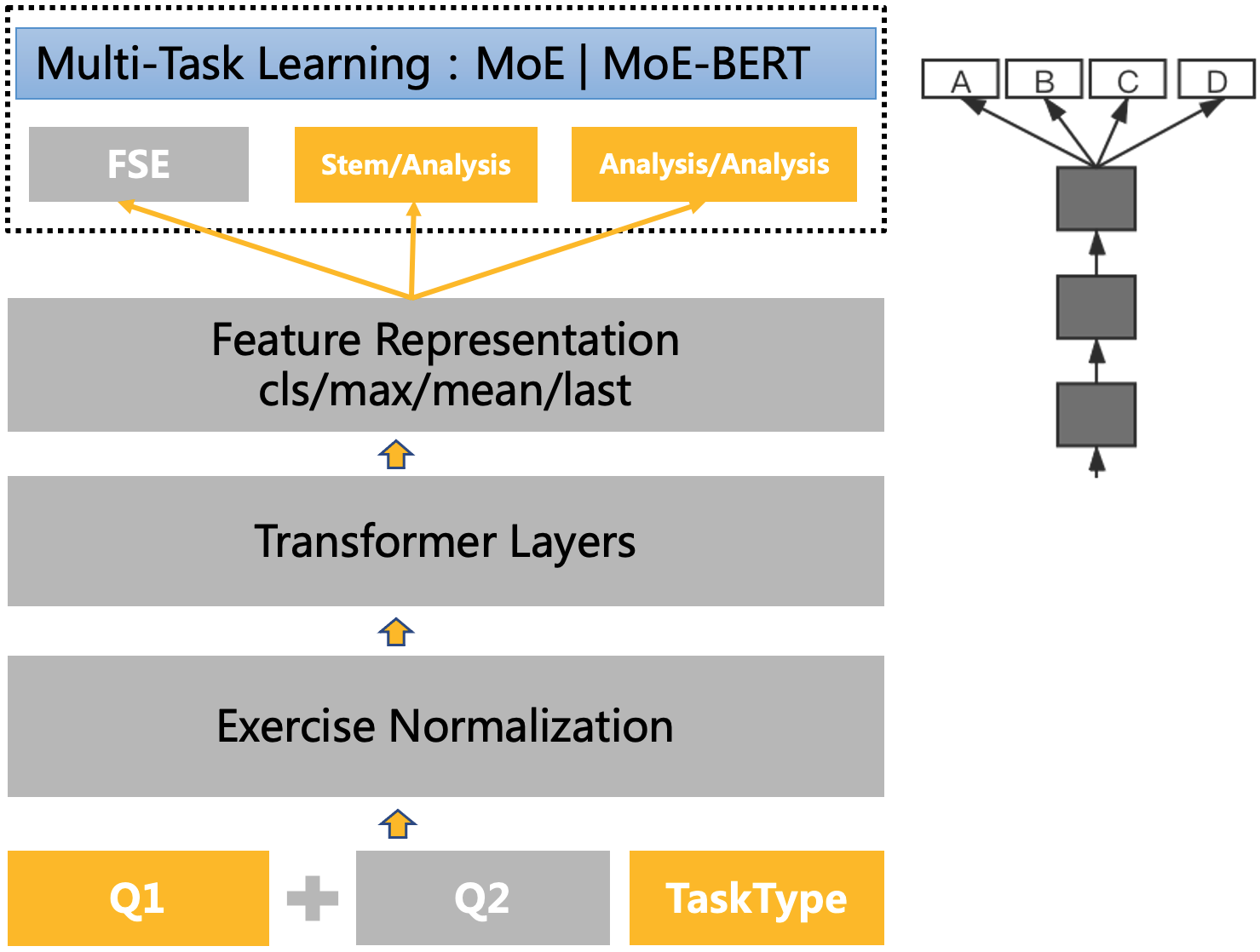}
\caption{The multi-task Model for FSE Task.}
\label{fig:f3}
\end{minipage}

\end{figure}

Here, we discover new auxiliary tasks T2 and T3 in an innovative way by using the structural attribute of the exercise in Figure-\ref{fig:f4}. 

Taking a labeled tuple <Exercise A, Exercise B, label> as an example, the stem and options of exercise A and exercise B constitute the FSE task T1. Exercise analysis A and exercise analysis B constitute a new task T2. The stem and exercise analysis can be constructed a new task T3+. Specifically, the positive pair can be directly constructed with the original exercise stem and exercise analysis, while the negative pair is slightly different that depends on the label of exercise pair. If the label is dis-similar, we can directly construct the exercise stem A and exercise analysis B. If the label is similar, directly construct may bring in noise. Here, we regard the current exercise stem and the same concepts of knowledge with randomly select an exercise analysis as a negative example(T3-).

\subsubsection{MoE enhanced Multi-Task Learning}

After discovering these two auxiliary tasks, we adopt the hard-sharing multitask learning same as \cite{liu2019multi} in Figure-\ref{fig:f3}.
As mentioned earlier, the exercise analysis which reflects the problem-solving ideas of exercise is very import to the FSE task. Introducing the multi-task learning, the model can learn more logical and semantic representation from exercise analysis and decrease the occurrence of over-fitting. The overall loss of multi-task is expressed as follows:
\begin{equation}
 \mathcal{L} = \alpha_{1}\cdot \mathcal{L}_{T1} + \alpha_{2} \cdot \mathcal{L}_{T2}+ \alpha_{3} \cdot \mathcal{L}_{T3}
\label{e1}
\end{equation}
where $\mathcal{L}_{T1}$ represents the loss of FSE task, $ \mathcal{L}_{T2}$ is the match loss of analysis, $ \mathcal{L}_{T3}$ is the match loss of stem and analysis,  $\alpha_i$ is the task coefficient which is match to $\sum_{i=1}^{n=3}(\alpha_i)=1$.

Since adjusting the task coefficient depends on manual attempts and expert experience, we adopt a 3-layer neural network to \textit{dynamically learn} the coefficients of tasks which is similar to MoE Layer\cite{shazeer2017outrageously,ma2018modeling} in information recommendation field. The MoE implementation is described in Appendix-\ref{ab}.

\subsection{Confidence Learning: Noise Dataset for Challenge 4}
\label{s41}
Due to the human labeled education data requires strong expertise, the differences exist in teachers' teaching experience and understanding of FSE problem. The statistical results show that the consistency rate of labeling samples is usually between 80-85\%, so there are some noises in the labeled samples. As we all know, if there is too much noise, the model will be easier to fit to the noise samples. Therefore, in order to prune or denoise the labeled samples and we introduce the confidence learning\cite{northcutt2021confident} to resolve this problem. 



\section{Experiments and Conclusions}
\label{s5}

In order to resolve the challenges we mentioned and verify the effectiveness of our solution, we compared with some benchmark methods(OVSM\cite{yu2014similarity}, BERT-Base\cite{devlin2018bert} and SBERT\cite{8992012}) on junior mathematics data set. We design the experiments and use the precision@k as the metric(k=1,3 and 5).

\begin{table}[ht!]
  \caption{Overall Performance of Results}
\centering
\tiny
    \begin{tabular}{l|rrr|lrrrc|c}
    \toprule
        Model & OVSM & BERT & SBERT & BERT$_{Edu}$ & +Norm & +MTL & +MoE & +CL(ExerciseBERT) & $\Delta$(vs. max(OVSM, BERT, SBERT))\\ \hline
        P@1 & 80.7\% & 75.8\% & \textbf{82.8\%} & 87.4\% & 88.6\% & 90.0\% & 91.2\% & \textbf{92.7\%} & \textbf{+9.9\%} \\ \hline
        P@3 & \textbf{79.6\%} & 74.2\% & 78.7\% & 83.9\% & 85.0\% & 88.2\% & 89.6\% & \textbf{90.6\%} & \textbf{+11.0\%} \\ \hline
        P@5 & \textbf{79.0\%} & 70.2\% & 75.9\% & 82.7\% & 83.8\% & 85.7\% & 86.7\% & \textbf{89.1\%} & \textbf{+10.1\%} \\\hline
    \end{tabular}
    \label{table:t1}
\end{table}

We summarize the experimental results in Table \ref{table:t1} and have the following observations:

\begin{itemize}
\item We can easily see that pre-trained methods(BERT$_{Edu}$) is able to boost the performance and consistently outperforms other baseline models. This proves that our model gains a better understanding of exercisers and is more efficiently to transfer from large unlabeled corpus to the label-scarce dataset;
\item The exercise normalization(+Norm) is useful to ensure identity and the experimental results confirmed this fact;
\item We discover new auxiliary tasks T2 and T3 in an innovative way by using exercise analysis. The experimental results also confirm the effectiveness of multi-task learning(+MTL) and provide more high level logic information for FSE task. In addition, the experimental results show that the MoE mechanism(+MoE) is very effective;
\item Confidence Learning(+CL)  aims to prune train-set and gains further improvement.
\end{itemize}





{
\small
\bibliography{sample-base}

\begin{thebibliography}{10}

\bibitem{chinese-bert-wwm}
Yiming Cui, Wanxiang Che, Ting Liu, Bing Qin, Ziqing Yang, Shijin Wang, and
  Guoping Hu.
\newblock Pre-training with whole word masking for chinese bert.
\newblock {\em arXiv preprint arXiv:1906.08101}, 2019.

\bibitem{devlin2018bert}
Jacob Devlin, Ming-Wei Chang, Kenton Lee, and Kristina Toutanova.
\newblock Bert: Pre-training of deep bidirectional transformers for language
  understanding.
\newblock {\em arXiv preprint arXiv:1810.04805}, 2018.

\bibitem{8992012}
M.~{Feng}, Y.~{Chen}, Y.~{Guo}, Y.~{Zhao}, and G.~{Fu}.
\newblock Learning text representations for finding similar exercises.
\newblock In {\em 2019 IEEE International Conference on Consumer Electronics -
  Taiwan (ICCE-TW)}, pages 1--2, 2019.

\bibitem{feng2019learning}
Mengfei Feng, Yishuai Chen, Yuchun Guo, Yongxiang Zhao, and Guowei Fu.
\newblock Learning text representations for finding similar exercises.
\newblock In {\em 2019 IEEE International Conference on Consumer
  Electronics-Taiwan (ICCE-TW)}, pages 1--2. IEEE, 2019.

\bibitem{liu2018finding}
Qi~Liu, Zai Huang, Zhenya Huang, Chuanren Liu, Enhong Chen, Yu~Su, and Guoping
  Hu.
\newblock Finding similar exercises in online education systems.
\newblock In {\em Proceedings of the 24th ACM SIGKDD International Conference
  on Knowledge Discovery \& Data Mining}, pages 1821--1830, 2018.

\bibitem{liu2019multi}
Xiaodong Liu, Pengcheng He, Weizhu Chen, and Jianfeng Gao.
\newblock Multi-task deep neural networks for natural language understanding.
\newblock {\em arXiv preprint arXiv:1901.11504}, 2019.

\bibitem{ma2018modeling}
Jiaqi Ma, Zhe Zhao, Xinyang Yi, Jilin Chen, Lichan Hong, and Ed~H Chi.
\newblock Modeling task relationships in multi-task learning with multi-gate
  mixture-of-experts.
\newblock In {\em Proceedings of the 24th ACM SIGKDD International Conference
  on Knowledge Discovery \& Data Mining}, pages 1930--1939, 2018.

\bibitem{nazaretsky2018kappa}
Tanya Nazaretsky, Sara Hershkovitz, and Giora Alexandron.
\newblock Kappa learning: A new method for measuring similarity between
  educational items using performance data.
\newblock {\em arXiv preprint arXiv:1812.08390}, 2018.

\bibitem{northcutt2021confident}
Curtis Northcutt, Lu~Jiang, and Isaac Chuang.
\newblock Confident learning: Estimating uncertainty in dataset labels.
\newblock {\em Journal of Artificial Intelligence Research}, 70:1373--1411,
  2021.

\bibitem{antlr4}
Terence Parr and Sam Harwell.
\newblock Antlr (another tool for language recognition).
\newblock \url{https://github.com/antlr/antlr4}.

\bibitem{pelanek2019measuring}
Radek Pel{\'a}nek.
\newblock Measuring similarity of educational items: An overview.
\newblock {\em IEEE Transactions on Learning Technologies}, 2019.

\bibitem{shazeer2017outrageously}
Noam Shazeer, Azalia Mirhoseini, Krzysztof Maziarz, Andy Davis, Quoc Le,
  Geoffrey Hinton, and Jeff Dean.
\newblock Outrageously large neural networks: The sparsely-gated
  mixture-of-experts layer.
\newblock {\em arXiv preprint arXiv:1701.06538}, 2017.

\bibitem{yin2019quesnet}
Yu~Yin, Qi~Liu, Zhenya Huang, Enhong Chen, Wei Tong, Shijin Wang, and Yu~Su.
\newblock Quesnet: A unified representation for heterogeneous test questions.
\newblock In {\em Proceedings of the 25th ACM SIGKDD International Conference
  on Knowledge Discovery \& Data Mining}, pages 1328--1336, 2019.

\bibitem{yu2014similarity}
Jing Yu, Dongmei Li, Jiajia Hou, Ying Liu, and Zhaoying Yang.
\newblock Similarity measure of test questions based on ontology and vsm.
\newblock {\em The Open Automation and Control Systems Journal}, 6(1), 2014.

\end{thebibliography}
}


\appendix
\section{Appendix}

\subsection{Data Sets.}
\label{s511}
The Tencent education platform\footnote{Tencent education platform: https://edu.tencent.com} contains millions of exercises and we only choose about 350K junior math exercises for our experiments. We sample 1.5K seed exercises and construct 23K exercise pairs through the BM25 match and some strategies such as random choose and random with concept. Then theses exercises are labeled with several similar exercises and each given exercise is labeled by three teachers. We choose the majority numbers of votes as the label for the similar exercise. We split our data set randomly via the seed exercises into three parts: 80\% is for training set, 10\% is for validation set and 10\% is for test set. Finally, we only report the performances on the test set.
\subsection{Experimental Settup}
\label{s522}
We implement all the models with Tensorflow in our experiments. In the pre-training stage, we pre-train our models with MLM objective, continuing from the published checkpoint, BERT-base-chinese. We pre-train our model 
for 200K steps, and the first 3000 steps are for warm-up. The rest of the hyper-parameters are the same as BERT-base. In the fine-tuning stage, we train our model in the multi-task paradigm. The multi task module adopts three layers of neural network, and the output sizes of hidden layer of each layer are 768,768 and 3. We apply the Adam method to optimize our model. The learning rate is $2e-5$, the number of training epoch is 3. We conduct our experiments with 2 Tesla T4 GPUs.

\section{Implementation of MoE Layer}
\label{ab}
We adopt a 3-layer neural network to dynamically learn the coefficients of tasks which is similar to MoE Layer\cite{ma2018modeling} in the information recommendation field. The detail operations are as follows:
First of all, we concat the feature representations of the different tasks:
\begin{equation}
feature= concat(Fe_{T_{1}}, Fe_{T_{2}}, Fe_{T_{3}})
\label{e2}
\end{equation}
Secondly, for the feature representation, we learn the parameter coefficients through a three-layer neural network.
\begin{equation}
    \alpha = (\alpha_{1},\alpha_{2},\alpha_{3}) = concat(Gate_1{(Fe_{T_{1}})}, Gate_2{(Fe_{T_{2}})}, Gate_3{(Fe_{T_{3}})})  
\label{e3}
\end{equation}
where $Gate_i(.)$ is the $i$-th expert network with a three-layer neural network.

Finally, we assign different task weights to different tasks as the above  Formula-\ref{e1}.

\section{Problem Formulation}
\label{s30}
As mentioned earlier, similar exercises are those having the same purpose which is related with the semantics of exercises.

\newtheorem{definition}{Definition}
\begin{definition}
Given a set of exercises including stem, option, concept of knowledge and exercise analysis, our target is to learn a model $\mathcal{F}$ which can be used to measure the similarity scores pairs and find similar exercises for any exercise E by ranking the candidate ones $\mathcal{D}$ with similarity scores:
\begin{equation}
    \mathcal{F}(E, \mathcal{D}, \Theta) \to \mathcal{R}^{s}
\end{equation}
where $\Theta$ is the parameters of $\mathcal{F}$, $\mathcal{D}=(E_1, E_2, E_3, \cdots)$ are the candidate exercises for $E$ and $\mathcal{R}^{s} = (E_1^s, E_2^s, E_3^s, \cdots)$ are the candidates ranked in descending order with their similarity scores $(S(E,E_1^s),S(E,E_2^s),S(E,E_3^s),\cdots)$. The similar exercises for E are those candidates having the largest similarity score.
\end{definition}

\section{Testing}
After obtaining the trained ExerciseBERT, for any \textit{exercise E} in the testing stage, we could find its similar exercises by ranking the candidate ones according to their similarity scores, and finally return the accurate Top-K similar exercises. We use the Precision@K as the metric. Precision@K is calculated as follows:
\begin{equation}
Precision@k = \sum_{i=1}^{N}\frac{true \ positives \ @ k}{(true \ positives \ @ k) + (false \ positives \ @ k)}
\end{equation}
where k=1, 3 and 5 and N is the number of seed exercises.

\section{Visualization Analysis}

\begin{figure}[!htbp]
\centering
\subfigure[Group 1]{
\includegraphics[width=5.0cm]{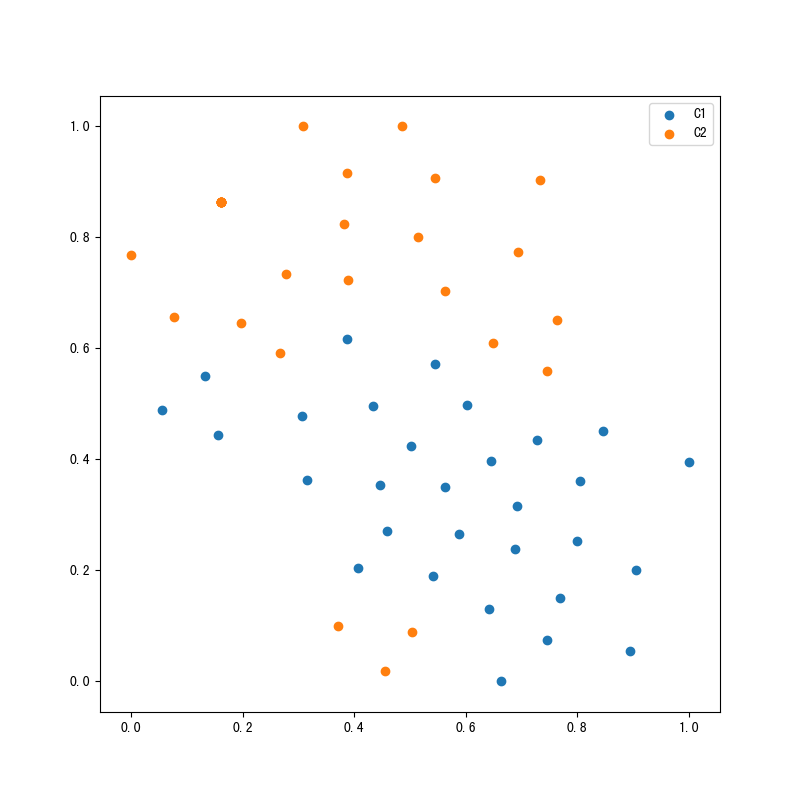}
\label{fig:a}
}
\quad
\subfigure[Group 2]{
\includegraphics[width=5.0cm]{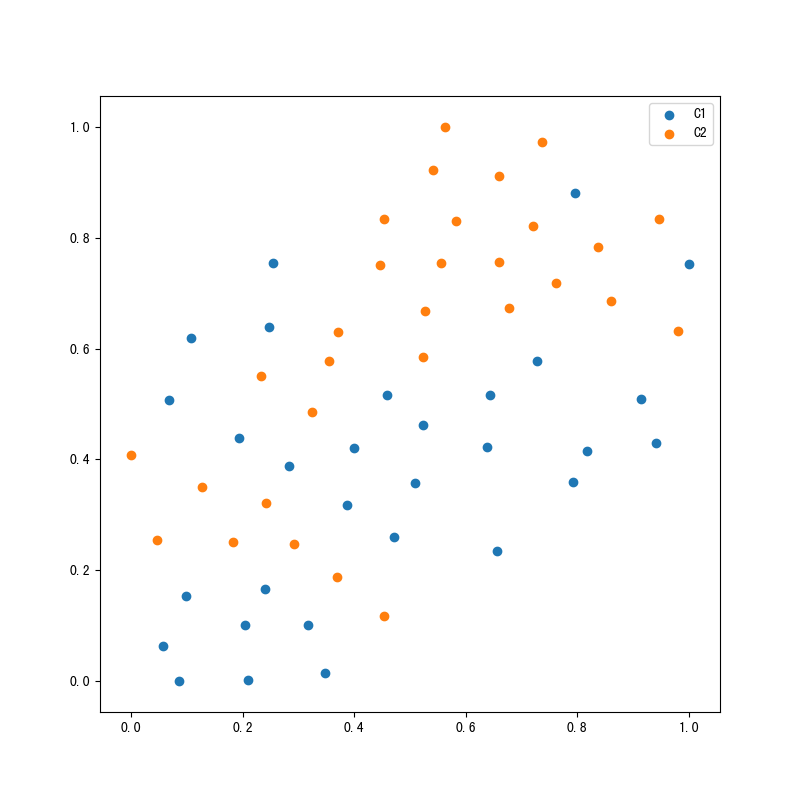}
\label{fig:b}
}
\quad
\subfigure[Group 3]{
\includegraphics[width=5.0cm]{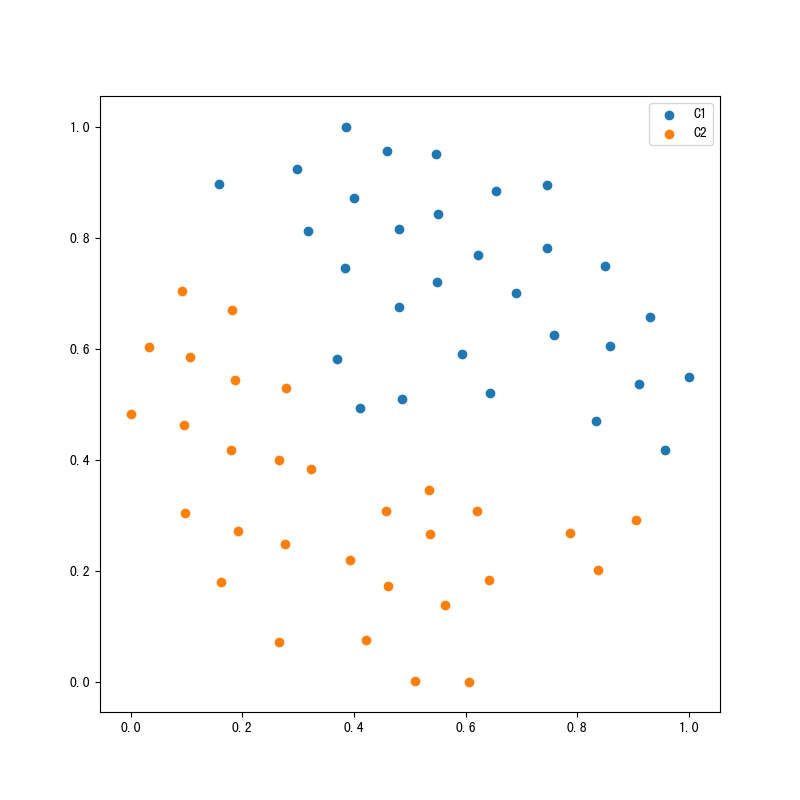}
\label{fig:c}
}
\quad
\subfigure[Group 4]{
\includegraphics[width=5.0cm]{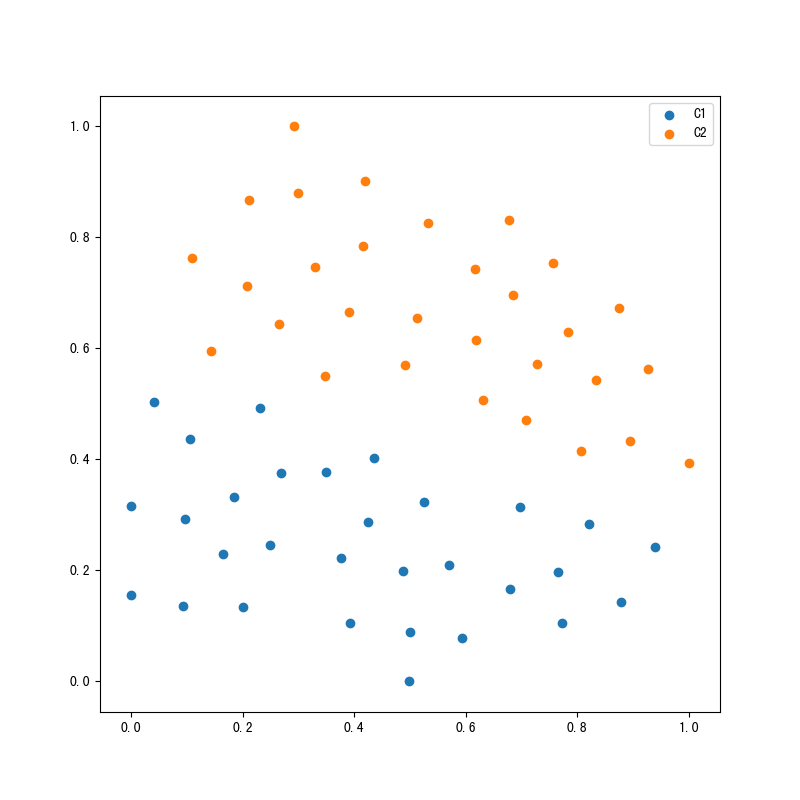}
\label{fig:d}
}
\caption{Visualization Of ExerciseBERT Representation}
\label{fig:vis}
\end{figure}

We conduct visualization analysis of the ExerciseBERT’s representation. It is important to learn exercise representations in which similar exercises are closer while dissimilar exercises are farther. To show the results intuitively, we first select four groups of exercises under two concepts that are randomly selected(the first two groups have the close knowledge concepts while the latter two are different), and then reduce the dimension of obtained representations by t-SNE. The visualization results are shown in Figure-\ref{fig:vis} and we can get two interesting phenomena. On the one hand, if the knowledge concepts of C1 and C2 are relatively similar (Fig-\ref{fig:a} and Fig-\ref{fig:b}), their exercises representation are also relatively close. There may be some overlapping parts because similar exercises are those having a similar logic and relationships including not only knowledge concept but also some other information such as problem-solving ideas. On the other hand, if the knowledge concepts of C1 and C2 are quite different (Fig-\ref{fig:c} and Fig-\ref{fig:d}), they are unlikely to become similar exercises and their exercises representation have a big difference. Thus, the results show that ExerciseBERT has a good exercise representation.

\end{document}